\documentclass{article}
\usepackage{spconf,amsmath,epsfig,caption}
\usepackage[utf8x]{inputenc}
\usepackage{url}

\usepackage[colorinlistoftodos]{todonotes}

\title{Deep learning for semantic segmentation of remote sensing images with rich spectral content}
\name{A. Ben Hamida\textsuperscript{1,2},
	  A. Benoit\textsuperscript{1},
      P. Lambert\textsuperscript{1},
      L. Klein\textsuperscript{1},
      C. Ben Amar\textsuperscript{2},
      N. Audebert\textsuperscript{3,4},
      S. Lefèvre\textsuperscript{4}}
\address{\textsuperscript{1} Univ. Savoie Mont Blanc, LISTIC, F-74000 Annecy, France
~~\\(amina.ben-hamida,alexandre.benoit, patrick.lambert, louis.klein)@univ-smb.fr
~~\\
\textsuperscript{2}
REGIM, Ecole Nationale d’Ingénieurs de Sfax, 3038 Sfax, Tunisie
~~\\chokri.benamar@ieee.org\\
\textsuperscript{3}
ONERA, \textit{The French Aerospace Lab}, 91761 Palaiseau, France\\
\textsuperscript{4}
Univ. Bretagne-Sud, UMR 6074, IRISA, 56000 Vannes, France
~~\\(nicolas.audebert,sebastien.lefevre)@irisa.fr
}
\begin{document}
%
\maketitle
\begin{abstract}
With the rapid development of Remote Sensing acquisition techniques, there is a need to scale and improve processing tools to cope with the observed increase of both data volume and richness. Among popular techniques in remote sensing, Deep Learning gains increasing interest but depends on the quality of the training data. Therefore, this paper presents recent Deep Learning approaches for fine or coarse land cover semantic segmentation estimation.  Various 2D architectures are tested and a new 3D model is introduced in order to jointly process the spatial and spectral dimensions of the data. Such a set of networks enables the comparison of the different spectral fusion schemes. Besides, we also assess the use of a ``noisy ground truth'' (i.e. outdated and low spatial resolution labels) for training and testing the networks.
\end{abstract}

\section{Introduction}
Automated land cover mapping based on satellite image analysis and classification is a well-known challenge which is of a great interest for many fields, such as agriculture \cite{agriculture} and risk monitoring \cite{risk}. The recently launched Sentinel-2 satellite constellation provides a richer content in spatial, spectral and temporal domains, and produces a huge amount of images to process daily. In this context, Deep Learning (DL) appears as an appealing alternative to traditional \emph{shallow} classification approaches to deal with such a massive amount of data. A DL architecture is a deep artificial neural network composed of a hierarchical succession of neuron layers performing linear and non-linear processing. Mimicking the human brain behavior, the network tuning (typically millions of parameters) is automatically performed thanks to a supervised training process on large datasets that are generally associated to some ``ground truth'' knowledge. However, the ground truth quality is essential to reach satisfying performances.

Some recent works already use deep neural nets to process remotely sensed images. In \cite{PVZ:16}, the authors compared different well-established deep architectures (AlexNet, AlexNet-small, VGG) for the classification of SAT-4/SAT-6 dataset (from US National Agriculture Imagery Program, NAIP) using Convolutional Neural Networks (CNN). Pirotti et al.~\cite{2016_BenchMark_Pirotti} benchmarked different machine learning methods (including multilayered perceptron) for classification of Sentinel-2 images. In \cite{Bids}, we proposed a new 3D CNN architecture for hyperspectral data pixelwise classification (semantic segmentation). In \cite{dcnn2,dcnn3}, we adapted the SegNet architecture to achieve semantic segmentation of multimodal airborne imagery.

In this paper, we rely on the recent DenseNet~\cite{dense} and SegNet~\cite{SegNet} architectures to perform land cover semantic segmentation of large multispectral Sentinel-2 images. These architectures are experimentally assessed through two different use cases, namely fine and coarse resolution estimation. We also introduce a new 3D DenseNet network in order to jointly process both spatial and spectral dimensions. In addition, we suggest the use of a ``noisy ground truth'' (i.e. outdated and low spatial resolution labels) for both training and testing. The idea is to explore the feasibility of outdated low quality knowledge integration for modern image sensors and analysis methods. Experiments are conducted in a wide region between France, Switzerland and Italy relying on the reference areas of GlobCover (ESA 2009 Global Land Cover Map) annotation and 2016 ESA Sentinel-2 images.

\section{Spectral channel fusion and Deep Neural Networks}\label{sec:soa}
When dealing with multispectral images, strategies for processing and fusing spectral channels are numerous. Deep neural networks enable a large variety of choices from the component operator to architectural levels. At the low component level, the classical approach is to use 2D convolution layers from the first stages of the network. In such a configuration, each neuron applies a specific filter to each channel and then fuses (sums) the resulting maps. Such an approach enables the early combination of each channel of multispectral image sensors whatever their wavelength. One can also rely on 3D convolution neurons that consider input images as 3D volumes. These volumes are then filtered by 3D filters that locally combine the spatial and spectral information. With such a strategy, cascading 3D filters enables all the available wavelength bands to be combined in a growing bandwidth manner. From a coarse network architecture point of view, the advances in neural network modeling led to an increase in architecture depth and a variety of neural layers combination enabling data representation levels to be also fused together.

In classical approaches such as AlexNet~\cite{krizhevsky2012}, a neuron layer is only fed by the previous layer output such that low-level representations are not directly included at the decision level that occurs in the last layer. However, recent ideas enable a given neuron layer to fuse the information coming from many more previous neural layers. The Residual Network (\textbf{ResNet}) components \cite{ResNetwork} consist in the fusion (pixelwise sum) between the resulting representation of a block of 2D convolution filters and its input. This enables to mix input data with a new slightly increased representation level.

One step further, the \textbf{SegNet} architecture \cite{SegNet} for semantic segmentation adopts an auto-encoder structure, with an encoder that models the data at multiple scales and a decoder that up-samples the internal representation and projects it into the mapping space. To do so, the location of the locally maximal activation from the lower layers are fed forward directly to their up-sampling counterpart in the last layers, which allows the decoder to relocate abstract features into the most salient points originally detected. This information is actually crucial for accurate semantic map boundary reconstruction.

Even further, the recent \textbf{DenseNet} architecture \cite{dense} proposes intensive layer fusion. Given a $Dense Block$ that consists in a set of convolution layers working at the same scale, each neural layer processes the concatenation of all its previous layers thus enabling the fusion of very numerous representation levels. Similarly to SegNet, its semantic segmentation extension \cite{densetiramisu16} adds a decoding path to generate the semantic map. However, on the decoding branch, the fusion not only consists in intra dense block layers fusion but it also relies, at the input of each decoding block, on the concatenation of the preceding high level feature maps and the ones coming from the encoding block at the same scale. The strength of such an approach is the intensive feature map reuse all along the network thus enabling the number of parameters to be significantly reduced while increasing performance.

\section{Proposed approaches}
In this paper we propose to explore the capabilities of some recent neural networks discussed in Sec.~\ref{sec:soa} for land cover mapping.
Conversely to public datasets provided with recent contests such as the  semantic labeling benchmark over ISPRS Vaihingen and Potsdam cities that relies on high-quality ground truth, we propose to train deep networks on a wide area, relying on coarse and noisy reference data. This realistic scenario allows to exploit previous knowledge such as outdated and coarse resolution land cover maps. These maps bring both confident annotations on qualitative and temporally stable regions, while presenting erroneous values on area boundaries but also on areas where changes occurred before image acquisition.
In this context, training neural networks in a supervised way is a delicate step and we propose two different use cases considering the following architectures, whose results are reported in Tab.\ref{tab:deepnets}.
\begin{figure}[tp]
  \centering
  \includegraphics[width=8cm, height=6cm]{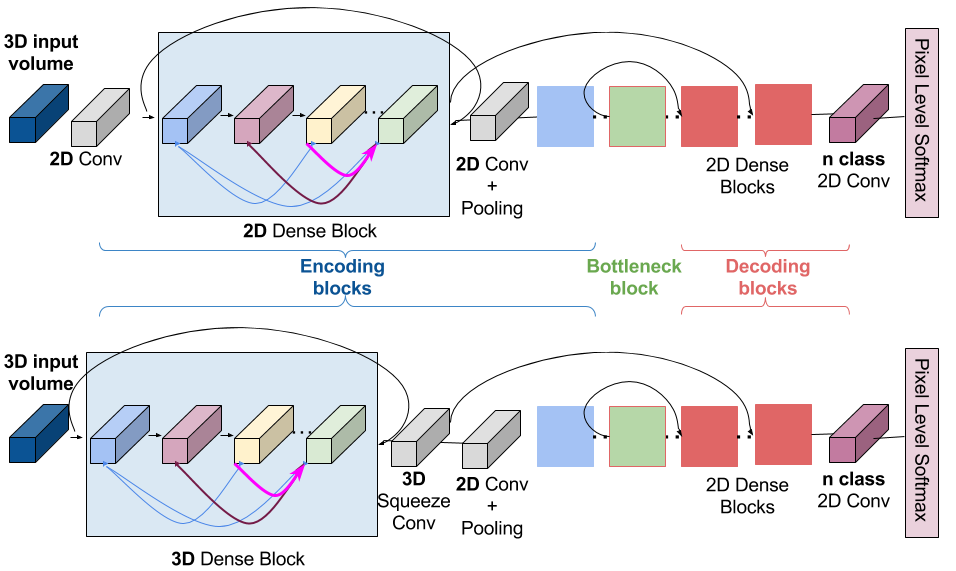}
  \caption{\textbf{Top:} reference DenseNet architectures. \textbf{Bottom:} our proposed 3D DenseNet }
  \label{fig:dense3d}
  \vspace{-0.5cm}
\end{figure}

First, we consider fine-grained pixel-level land cover estimation using very light architectures with few parameters using DenseNet~\cite{dense}. 3 basic configurations are proposed, with either 2 or 3 dense blocks at the encoder and decoder levels, different number of layers per block and either 12 or 16 neurons per layer. As shown in Fig.~\ref{fig:dense3d}, an extension of this architecture is proposed by introducing 3D convolution neurons in the first dense block. As already proposed in \cite{Bids}, we aim to early combine spectral channels but only for adjacent bands. To do so, the 48$\times$2D convolution neurons input layer of the reference model is removed thus enabling all the input spectral bands to feed a new first dense block exclusively composed of 3D convolutions with kernel size 3$\times$3$\times$3 using 4 layers of 16 neurons each.
Then, if 9 spectral bands regularly sampled along the wavelength dimension are chosen, the neurons of the 4th layer of the 3D dense block have processed data that depend on the full spectral bandwidth. An intermediate 3D convolution is then added next to ensure the 3D to 2D data transition by squeezing the spectral dimension.
Compared to 2D convolution based models, the overall number of parameters decreases by at least 10\%.


Second, we consider land cover estimation at the coarse ground truth scale using a very large architecture derived from SegNet~\cite{SegNet}. We modify the original SegNet by allowing the input to have more than 3 bands by enlarging the first convolution layer. The encoder is based on VGG-16~\cite{details}, which alternates blocks of 2 or 3 convolutions with $3\times3$ kernels with max-pooling layer that downsamples the data by a factor of two. The decoder is symmetric to the encoder, with sparse $2\times$ up-sampling instead of pooling and convolutions to densify the activations. The up-sampling is performed using the locations of the maxima found during pooling in the encoder, that are skipped through the network. After each block in the decoder, a $1\times1$ convolution layer projects the feature maps into the label space. We add decoding blocks until the estimation has a higher resolution than the ground truth. The multiscale estimations are then interpolated at full scale and averaged for loss computation. The gradient is then back-propagated in a deeply supervised fashion, to enforce a better multiscale spatial regularity. The full architecture has more than 27M parameters and is the largest model presented in this work.

\begin{table*}[]
\centering
\caption{Description of the considered architectures and Overall Accuracy (OA) after reference class boundary erosion of 200m. For DenseNet, e[x,y], d[y,x] and b[z] list the number of blocks and their respective number of layers for respectively the encoding and symmetric decoding branches and the bottleneck block. The number of neurons per layer is parametrized by $g$.}
\label{tab:deepnets}
\begin{tabular}{|llrlllll|}
\hline
Use Case & Models & \#params & \#scales & \multicolumn{2}{c}{OA/9 bands [B1-B8a]} & \multicolumn{2}{c|}{OA/13 bands}\\
& &  & & D1 & D2& D1 & D2\\
\hline \hline
Fine-grained estimation & DenseNet & & & &&&\\
&e[2,3],b[4],d[3,2] g12 & 0.23M  & 3 & 30.1\% & 57.9\% & 35.9\% &51.8\%\\ 
&e[4,5],b[7],d[5,4] g16 & 1.00M & 3 &  29.5\% & 55.3\% & 27.2\% & 55.4\% \\ 
&e[4,4,4],b[4],d[4,4,4] g16  & 1.08M & 4 & 25.7\%& 52.3\%& 28.5\%& 51.4\%\\ 
\hline
Fine-grained estimation &3D DenseNet & && &  &  &\\

&$e[4,5]_{3D}$,b[7],d[5,4] g16 & 0.88M & 3 &  25.1\% & 39.5\% &- &-\\ 
&$e[4,4,4]_{3D}$,b[4],d[4,4,4] g16  & 0.92M & 4 & 26.5\%& 41.2\%&- &-\\ 
\hline \hline
Coarse estimation&SegNet & 27.00M & 5 & - & - & 66.9\% & 83.9\%\\
\hline
\end{tabular}
\vspace{-0.5cm}
\end{table*}

\section{Experiments}
\begin{figure}[tp]
  \centering
  \includegraphics[width=8cm, height=4cm]{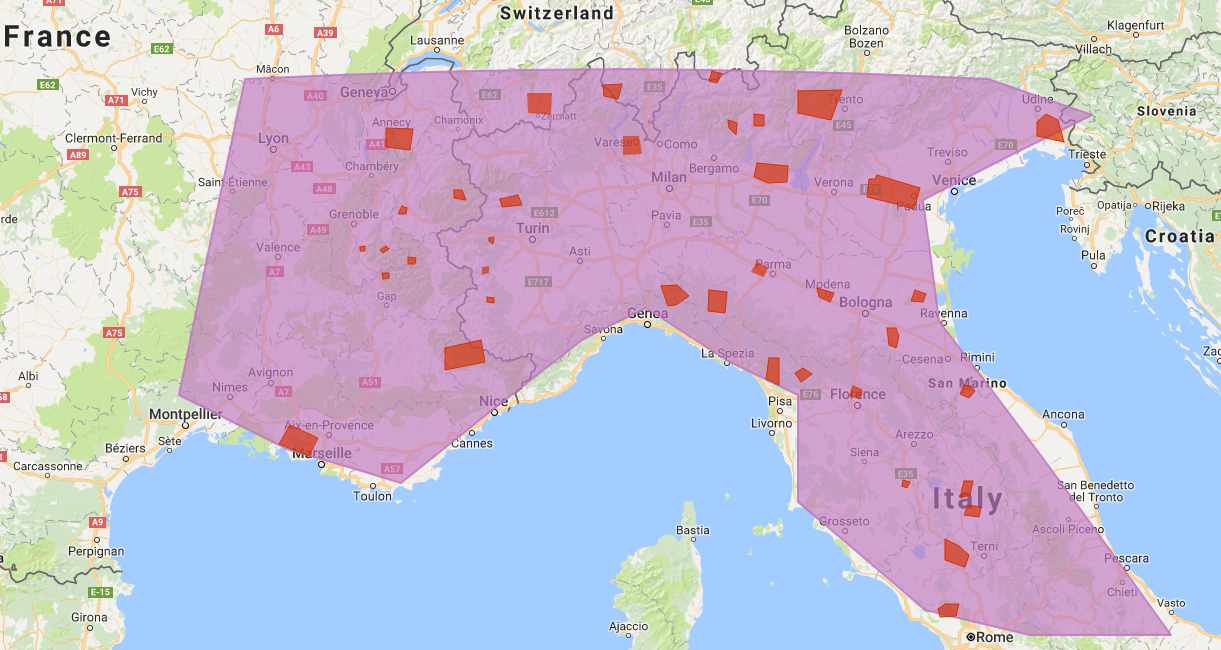}
  \caption{Experimental datasets, training images were extracted from the overall pink region while test regions are the small orange ones. Each image covers GlobCover reference areas.}
  \label{fig:dataset_overview}
\vspace{-0.4cm}
\end{figure}

Our experiments have been made possible thanks to the MUST computing center of the University of Savoie Mont Blanc.
The proposed neural networks are trained and evaluated on a dataset extracted from a region between France, Switzerland and Italy as shown in Fig.~\ref{fig:dataset_overview}. Images have been acquired by the Sentinel-2 sensor within the May-October 2016 period, while land cover ground truth is the 2009 GlobCover map $ESA/GLOBCOVER\_L4\_200901\_200912\_V2\_3$. The Google Earth Engine \cite{googleearthengine} was used to extract Sentinel Images but only on the reference areas of GlobCover. Some subregions of this dataset have been put apart to serve as the test dataset. Two datasets are proposed and detailed in Tab.\ref{tab:datasets}: the first one covers all the May-October period and does not include areas with clouds (opaque and cirrus). The second one only covers the summer period but includes clouds so that they can be considered as an additional class to detect. Both datasets are then multitemporal with frequent observations of the same areas at different timestamps.

The Sentinel-2 data is interpolated to 20m/px resolution, while the GlobCover ground truth is at 300m/px resolution. The proposed networks are trained considering two strategies: the light ones are optimized at the image resolution level whatever the ground truth resolution is (\textit{i.e.} the models are trained against an interpolated ground truth), thus trying to estimate at the scale of the input even if the reference is too coarse. The largest network (SegNet) however estimates multiple maps at several scales that are averaged and interpolated to the ground truth resolution. Training at a lower resolution alleviate errors along class borders in the GlobCover data.

\begin{table}[]
\centering
\caption{Considered datasets: images including clouds detected by the Sentinel-2 A60 additional band are excluded from the long period dataset (D1). The short period dataset (D2) includes clouds and enables training for their detection. Both datasets collect more than 150M pixels each.}
\label{tab:datasets}
\begin{tabular}{|llll|}
\hline
Dataset&\multicolumn{3}{c|}{\#images}\\
(period)  & train & test  & \#class \\
\hline \hline
D1, Long period no clouds & & & \\
 (May-Oct. 2016)& 140 & 54 & 23 \\
\hline \hline
D2, Short period with clouds & & & \\
(June-August 2016) & 158 & 39 & 24 \\
\hline
\end{tabular}
\vspace{-0.5cm}
\end{table}
As shown on Tab.~\ref{tab:deepnets}, estimation at a higher resolution than the reference provides an overall accuracy ranging between 25.1\% and 30.0\% on dataset D1 and 39.5\% and 57.9\% on D2. However, the overall accuracy metric introduces a discrepancy that penalizes the reported values. The metric is actually confident on stable and large areas but is less confident on heterogeneous areas. In such a noisy training context, resembling classes such as ``Rainfed croplands'' and ``Mosaic Cropland/Vegetation'' are often confused. However, these network are expressive enough to well detect classes such as ``Artificial surfaces'', ``Bare areas'', ``Water bodies'' and the added class ``Clouds''. When working with only the regularly sampled spectral bands (9 bands tests), 3D DenseNets can be applied. 3D approaches are performing worse but performance increases with model depth. Nevertheless, many more training images are required to train for the fine-grained approach.
However, coarse land cover estimation provides good results on both datasets. The large number of parameters of SegNet and the multiscale approach provides at least 66.9\% accuracy on the long period dataset that shows strong aspect changes of the same areas (D1) and 83.9\% on the more stable one (D2).
\section{Conclusion}
This paper reviews a variety of strategies provided by Deep Learning approaches to fuse the channels of multispectral sensors with the aim of land cover mapping. Optimizing such neural networks from a noisy land cover reference is challenging when dealing with reduced size datasets. Estimating at coarse image scale up to the one of the reference remains the most appropriate solution. Estimating at a higher resolution is difficult for the training step but also at the validation step to enable confident comparison. A first step forward would consist in enhancing the proposed models by enabling multiscale estimation for each of them. Going further, estimating at a finer resolution is challenging but required for land cover monitoring, and refined approaches should be investigated by taking advantage of the recently available mass of training data.

\bibliographystyle{IEEEbib}
\bibliography{refs}

\end{document}